\documentclass[hyphens]{article}
\usepackage[final]{neurips_2025}

\usepackage[utf8]{inputenc}
\usepackage[T1]{fontenc}
\usepackage{hyperref}
\hypersetup{
  colorlinks=true,
  linkcolor=MidnightBlue,
  citecolor=MidnightBlue,
  urlcolor=magenta,
}
\usepackage{url}
\usepackage{xurl}
\usepackage{booktabs}
\usepackage{amsfonts}
\usepackage{amsmath}
\usepackage{amssymb}
\usepackage{microtype}
\usepackage[table,dvipsnames]{xcolor}
\usepackage{xspace}
\usepackage{enumitem}
\usepackage{graphicx}
\usepackage{svg}
\usepackage{float}
\usepackage[numbers,sort&compress]{natbib}
\usepackage{tikz}
\usetikzlibrary{arrows.meta,positioning,fit}
\usepackage{makecell}
\usepackage{multicol,multirow}
\usepackage{threeparttable}
\usepackage{tabularx}
\usepackage{array}
\usepackage{ragged2e}
\usepackage[labelfont=bf]{caption}
\usepackage[section]{placeins}

\AtBeginEnvironment{thebibliography}{\raggedright\sloppy}
\usepackage{fancyhdr}
\newcolumntype{Y}{>{\RaggedRight\arraybackslash}X}
\newcolumntype{C}{>{\centering\arraybackslash}X}
\newcommand{\modelname}{Turing-20B-A2B\xspace}

\IfFileExists{figures/tensorboard.pdf}{}{%
  \immediate\write18{inkscape figures/tensorboard.jpg --export-filename=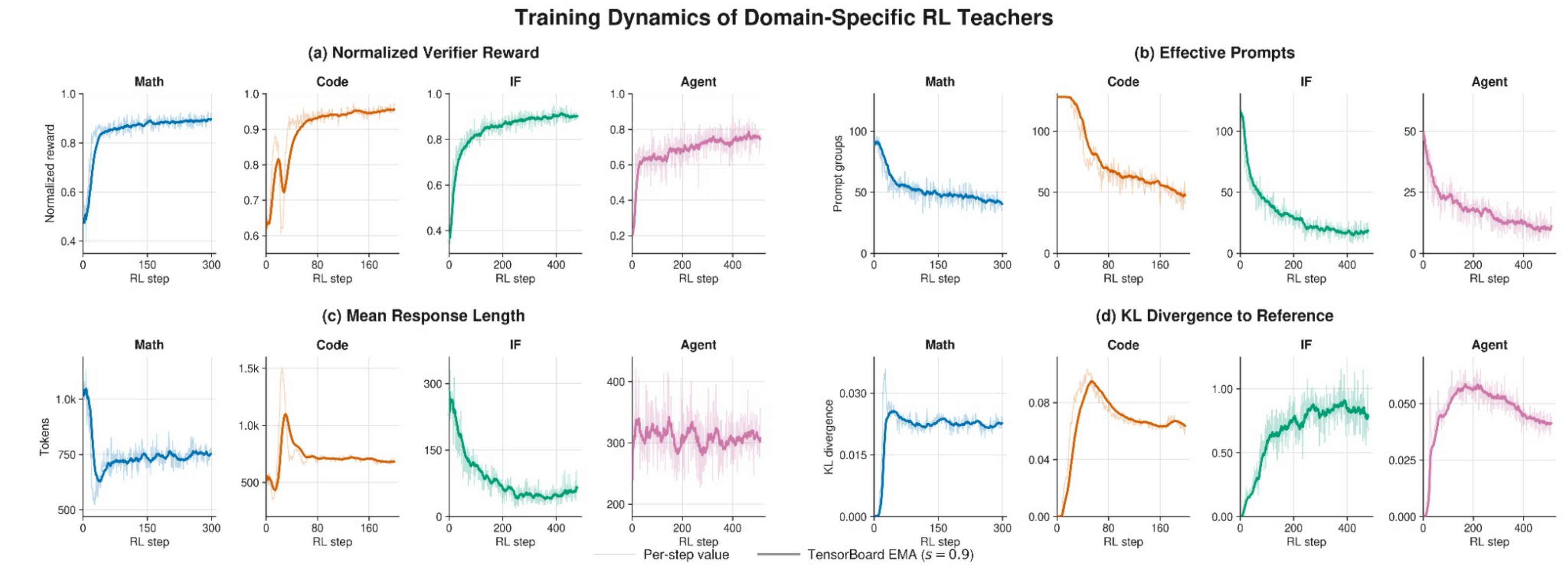}%
}
\IfFileExists{prism-uploads/image_pipeline_explicit_newline.pdf}{}{%
  \immediate\write18{inkscape prism-uploads/image_pipeline_explicit_newline.svg --export-filename=prism-uploads/image_pipeline_explicit_newline.pdf}%
}

\title{\fontsize{16}{18}\selectfont
Turing LLM Post-Training: Calibrated Distillation and Efficient Dynamic-MoE Execution}
\author{
  \includegraphics[height=0.8em]{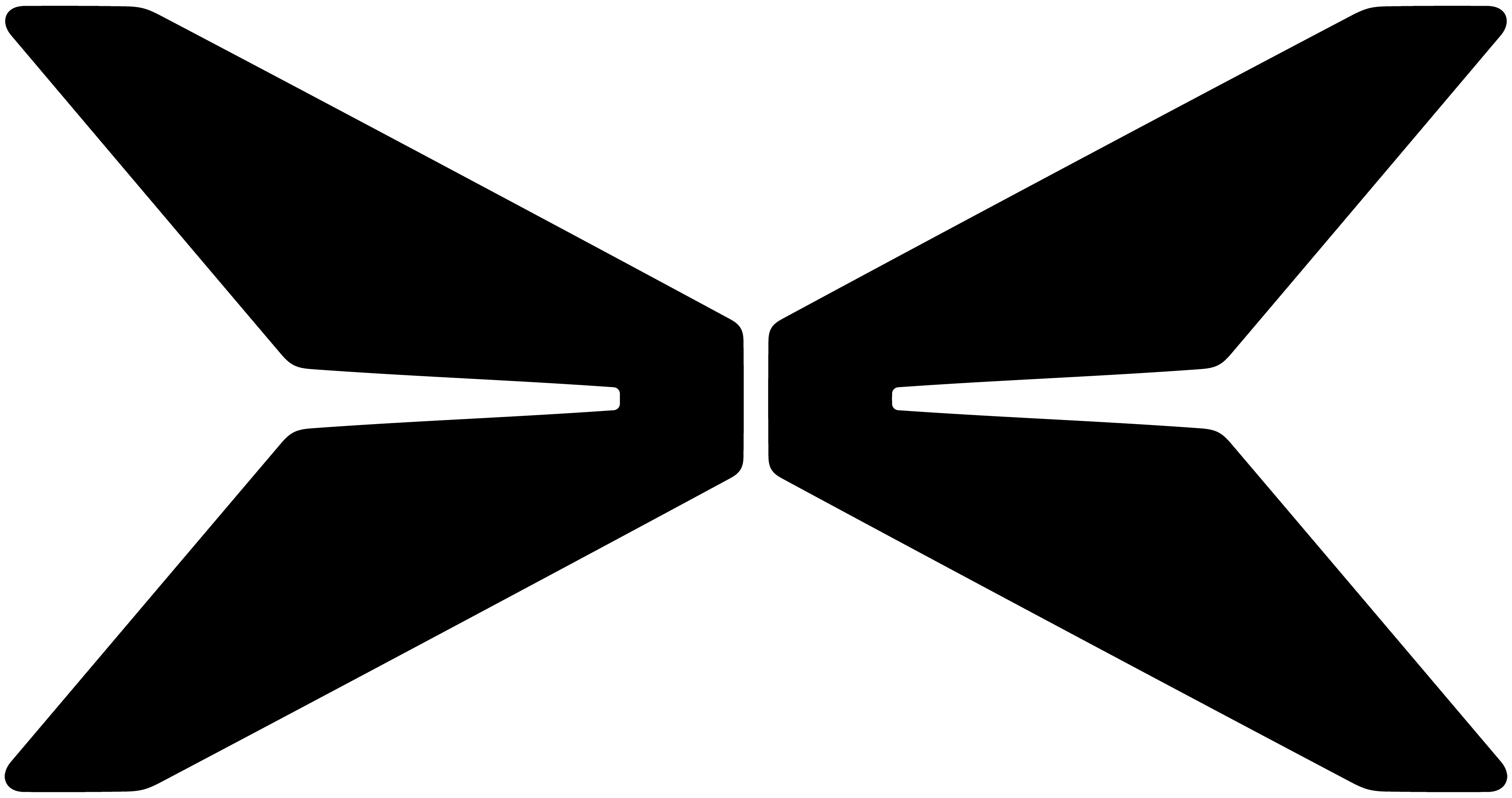}
  Foundation Model Team, XPeng Inc.\\
}

\begin{document}
\raggedbottom
\maketitle
\setlength{\headheight}{18pt}
\begingroup
\small
% !TeX root = ../main.tex
\vspace{-0.8em}
\begin{abstract}
We study post-training for dynamically routed mixture-of-experts (MoE) models, focusing on integrating independently optimized specialists while keeping consolidation efficient. Our approach combines calibrated multi-teacher on-policy distillation (C-MOPD), which matches each selected teacher on a local Top-$K$ support with teacher-wise loss normalization and clipping, with an execution path that uses post-capacity bounded-width dispatch, final-route replay, and asynchronous rollout--training overlap. On Turing-20B-A2B, the selected model achieves a four-domain normalized integration score of 0.9988 and retains at least 90.16\% of the measured specialist gain in every domain. At matched $K=32$, the full calibration objective scores 0.9962, compared with 0.9724 without calibration and 0.9834 without clipping. At a matched 8K response cap, the model exceeds both shared SFT and Qwen3-8B on all four reasoning-enabled mathematics and code benchmarks (unweighted mean: 85.92, versus 80.33 for Qwen3-8B and 74.21 for SFT), and improves over SFT on ten of twelve broader evaluation metrics. In a PPU-based profile, the combined execution configuration reduces consolidation step time from 398.7 to 139.3 seconds, an observed 2.86$\times$ speedup over serial execution. These results demonstrate strong specialist integration alongside more efficient consolidation, while reflecting the complete training and execution framework rather than any single component.
\end{abstract}

\noindent\textbf{Keywords:} LLM Post-Training, Mixture of Experts, On-Policy Distillation, Reinforcement Learning

% \noindent{\footnotesize\color{red}\textbf{Draft convention:} red text marks information that must be confirmed from author records or implementation before external release; it does not denote a request for additional experiments.}

\endgroup

\begingroup
\setlength{\intextsep}{6pt}
\begin{figure}[H]
  \centering
  \captionsetup{font=footnotesize,skip=4pt}
  \includegraphics[width=\textwidth]{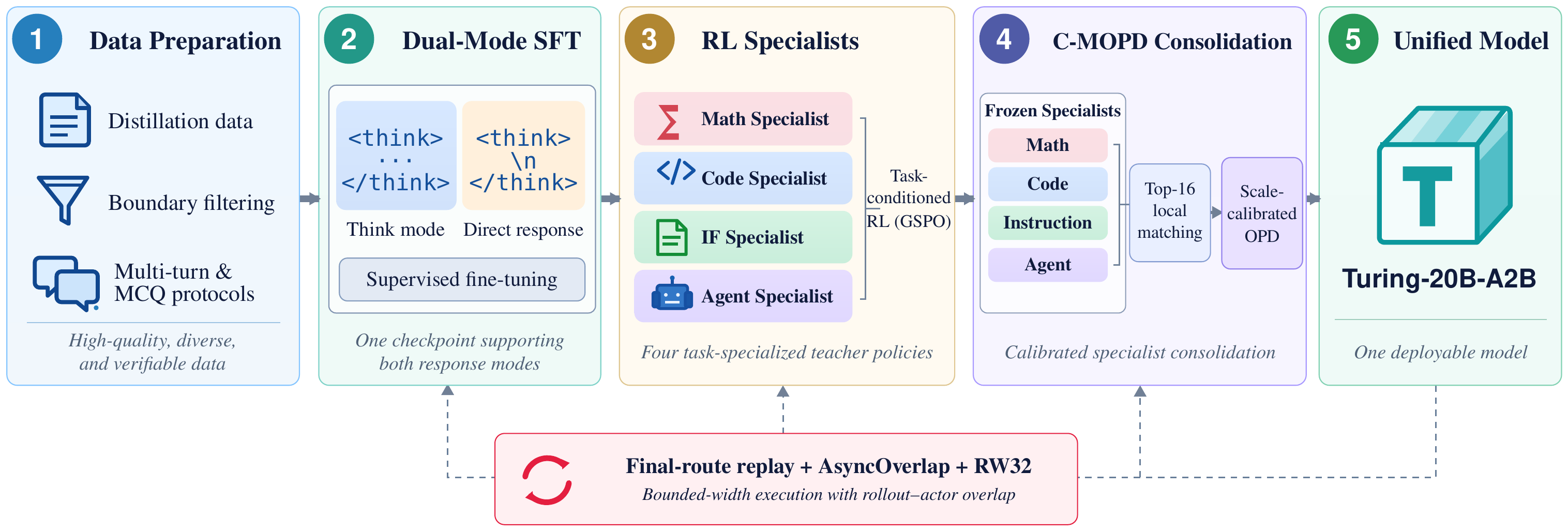}
  \caption{The Turing LLM post-training framework. Shared supervised initialization provides the starting point for domain-specialist RL. C-MOPD consolidates the frozen specialists into one student through teacher-local Top-$K$ matching and loss-scale calibration on student-generated prefixes. Bounded-width dispatch, route replay following R3, and asynchronous overlap support efficient dynamic-MoE execution in the consolidation workflow.}
  \label{fig:pipeline}
\end{figure}
\endgroup
\clearpage

% !TeX root = ../main.tex
\section{Introduction}
\label{sec:intro}

Modern LLM post-training increasingly relies on capability-specific optimization pipelines. Different capabilities may call for different data, feedback signals, training environments, and optimization recipes, allowing each pipeline to develop a strong specialist. Deploying these specialists separately can increase model-storage and serving overhead, motivating their consolidation into a single model with broader task coverage. We refer to the challenge of combining their complementary capabilities while retaining specialist gains as \emph{capability integration}.

Existing integration routes make different compromises. Mixed-domain reinforcement learning (RL) couples heterogeneous domains to a shared sampling schedule and optimization trajectory, while sequential RL can overwrite capabilities acquired in earlier stages. Off-policy distillation from fixed teacher trajectories can suffer from exposure mismatch, while parameter-space merging combines specialist checkpoints through weight-space operations. MOPD~\cite{mopd} offers a specialize-then-consolidate alternative: domain RL teachers are developed independently and then distilled into a student on the student's own rollouts.

Two practical challenges arise in our setting. First, token-level distillation losses can vary substantially in scale across teachers, complicating the use of shared optimization settings and fixed teacher weights. Second, the rollout--actor loop for a dynamically routed mixture-of-experts (MoE) student must handle variable-width expert dispatch and cross-engine synchronization while preserving route consistency between generation and training.

We study both challenges in the post-training of \modelname, a 20B-parameter dynamic MoE with approximately 2B active parameters per token~\cite{turingllm}. A shared SFT checkpoint initializes RL specialists for mathematics, code, instruction following, and agentic tool use. We then consolidate the frozen specialists into one student with calibrated multi-teacher on-policy distillation (C-MOPD). Following MOPD, task routing selects one teacher for each sample, and supervision is evaluated on student-generated prefixes. C-MOPD applies teacher-wise loss-scale normalization and token-level clipping to teacher-local Top-$K$ reverse-KL matching~\cite{revisitingopd}, reducing sensitivity to teacher-specific loss scales while retaining explicit teacher weights.

The consolidated student achieves a normalized four-domain integration score of 0.9988. With $K=16$ and $c_{\max}=5$, the selected model retains at least 90.16\% of the improvement achieved by the corresponding specialist over the shared SFT baseline in each domain. The final model produced by the specialist-training and consolidation pipeline also improves over the shared SFT baseline on all four mathematics and code benchmarks evaluated in reasoning mode and on ten of twelve metrics in the broader evaluation suite.

To improve consolidation efficiency, we combine asynchronous rollout--actor overlap with one-step policy lag, bounded-width dispatch after capacity filtering, and final-route replay following R3~\cite{r3}. With the post-capacity dispatch width capped at 32, the combined configuration reduces the profiled consolidation step time from 398.7 to 139.3 seconds, yielding a 2.86$\times$ speedup over serial execution.

Our contributions are:

\begin{itemize}[leftmargin=1.4em]

\item \textbf{Specialist capability integration.}
We instantiate an end-to-end specialize-then-consolidate pipeline for a dynamic MoE, connecting a shared SFT initialization, four verifier-driven RL specialists, and a single C-MOPD student.

\item \textbf{Calibrated multi-teacher distillation.}
We introduce C-MOPD, which combines teacher-local Top-$K$ reverse-KL matching with teacher-wise loss-scale normalization and token-level clipping under fixed teacher weights. In matched ablations at $K=32$, the full objective achieves a normalized integration score of 0.9962, compared with 0.9724 when normalization and clipping are both removed and 0.9834 when clipping alone is removed.

\item \textbf{Efficient dynamic-MoE execution.}
We adapt the consolidation workflow through bounded-width dispatch after capacity filtering, final-route replay, and asynchronous rollout--actor overlap to improve execution efficiency while maintaining cross-engine route consistency.

\end{itemize}

\autoref{sec:method} describes C-MOPD and the dynamic-MoE execution path.
\autoref{sec:experiments} presents capability-integration results, matched ablations, and execution profiles.
The appendices provide training recipes, complete results, diagnostic analyses, and reproducibility details.

% \input{sec/2_related_works}
% !TeX root = ../main.tex
\section{Method}
\label{sec:method}

\subsection{Framework Overview}

Our framework comprises three stages (\autoref{fig:pipeline}): shared supervised fine-tuning, domain-specific reinforcement learning, and multi-teacher on-policy distillation. Starting from a common SFT checkpoint, we train four specialists for mathematics, code, instruction following, and agentic tool use. C-MOPD then consolidates the frozen specialists into a single student using student-generated trajectories. We use \emph{specialist} to denote a complete domain teacher and \emph{expert} to denote an internal MoE module. Task-level teacher selection and token-level expert routing operate at different levels of the framework.

The execution workflow builds on the base model's capacity-constrained Quantile Routing~\cite{turingllm}. Post-capacity dispatch-width bounding limits the number of expert assignments represented and executed for each token. Final-route replay following R3~\cite{r3} preserves these discrete assignments during actor computation, while asynchronous overlap runs rollout generation concurrently with actor updates.

\subsection{Shared Initialization and Specialist Construction}
\label{sec:specialist-construction}

The shared SFT stage uses approximately 5.28M examples spanning mathematics, code, tool use, instruction following, long context, and dialogue.
The same model supports two prompt-selected response modes, following Qwen's convention~\cite{qwen3prompt}:
reasoning-enabled prompts end the assistant prefix with an open \texttt{<think>} marker, so generation proceeds in the reasoning block until \texttt{</think>}; direct-response prompts instead insert an empty \texttt{<think></think>} pair before the answer.
The data recipe includes teacher correction, capability-boundary filtering, and protocol-randomized multiple-choice construction; \autoref{app:training-recipe} describes these procedures and the data mixture.

Starting from this checkpoint, we train four domain specialists with GSPO, using 16 rollouts per prompt~\cite{gspo}. Each domain uses a task-specific prompt mixture and reward design, as summarized in \autoref{tab:task-rewards}. After training, we freeze the four specialists and use them as teachers for C-MOPD consolidation.

\begin{table}[t]
  \centering
  \caption{Prompt-form mixtures and reward components for specialist training. Think and Non-think denote reasoning-enabled and direct-response prompts; percentages give their sampling proportions within each domain.}
  \label{tab:task-rewards}
  \footnotesize
  \setlength{\tabcolsep}{4pt}
  \renewcommand{\arraystretch}{1.12}
  \begin{tabularx}{\textwidth}{@{}lccY@{}}
    \toprule
    \textbf{Task} & \textbf{Think} & \textbf{Non-think} & \textbf{Verification and reward} \\
    \midrule
    Code & 80\% & 20\% & Extraction, signature validity, compilation, and unit-test pass rate; auxiliary terms are gated by executable validity. \\
    Math & 80\% & 20\% & Final-answer correctness and normal completion, with penalties for truncation and repeated reasoning. \\
    IF & 20\% & 80\% & Constraint satisfaction or option correctness, with a preference for concise valid outputs. \\
    Agent & 100\% & 0\% & Tool-call format, argument correctness, task completion, and process feedback from a 20B process reward model. \\
    \bottomrule
  \end{tabularx}
\end{table}

\subsection{C-MOPD: Calibrated Multi-Teacher On-Policy Distillation}
\label{sec:mopd}

\subsubsection{Task-Conditioned On-Policy Supervision}

Following MOPD~\cite{mopd}, a task router selects one frozen domain teacher for each prompt $x_j$,
\begin{equation}
  d_j=r(x_j).
\end{equation}
In the synchronous formulation, the student samples a trajectory $y_j\sim\pi_\theta(\cdot\mid x_j)$. At each valid response position $t$, teacher $d_j$ is evaluated on the student-generated prefix $(x_j,y_{j,<t})$. Let $p_{\theta,j,t}$ and $q_{d_j,j,t}$ denote the student and selected-teacher next-token distributions on that prefix. When AsyncOverlap is enabled, trajectories are generated using the preceding student policy version, introducing a one-step policy lag (\autoref{sec:system}).

\subsubsection{Teacher-Local Reverse KL}

For each valid response position, define the selected teacher's local support
\begin{equation}
  S_{j,t}=\operatorname{Top}_{K}\!\left(q_{d_j,j,t}\right).
\end{equation}
We independently renormalize the student and teacher distributions within this support,
\begin{equation}
  \bar p_{\theta,j,t}^{S}(v)=
  \frac{p_{\theta,j,t}(v)}{\sum_{u\in S_{j,t}}p_{\theta,j,t}(u)},\qquad
  \bar q_{d_j,j,t}^{S}(v)=
  \frac{q_{d_j,j,t}(v)}{\sum_{u\in S_{j,t}}q_{d_j,j,t}(u)},
\end{equation}
and compute the token-level reverse KL divergence between the locally normalized distributions,
\begin{equation}
  \ell_{j,t}=D_{\mathrm{KL}}\!\left(
    \bar p_{\theta,j,t}^{S}\,\middle\|\,\bar q_{d_j,j,t}^{S}
  \right).
  \label{eq:topk-rkl}
\end{equation}
This local-support objective follows prior on-policy distillation work~\cite{revisitingopd}. It matches relative probabilities among the teacher's $K$ most probable tokens. Because both distributions are renormalized within $S_{j,t}$, the loss does not directly constrain the student's total probability mass outside this support. We compare the full objective with its calibration ablations at a fixed support size of K = 32 to assess the effects of loss-scale normalization and token-level clipping under this setting. We also evaluate different support sizes with the full objective at a clipping threshold of 5. The selected model uses K = 16, which achieves the highest observed macro integration score among the tested support sizes.

\subsubsection{Teacher-Wise Loss-Scale Calibration}

Token-level distillation losses may differ in scale across teachers. C-MOPD applies teacher-wise normalization and token-level clipping to reduce sensitivity to these differences.

Let $\mathcal{B}_\mu$ be a microbatch and $m_{j,t}\in\{0,1\}$ mark valid response tokens. For each teacher represented in the microbatch, we compute the standard deviation of its token losses,
\begin{equation}
  \sigma_d^{(\mu)}=\operatorname{Std}\!\left\{
    \ell_{j,t}:j\in\mathcal{B}_\mu,\ d_j=d,\ m_{j,t}=1
  \right\}.
\end{equation}
The scale statistic $\sigma_d^{(\mu)}$ is detached from the computation graph. With a scale floor of 0.1, the calibrated token loss is
\begin{equation}
  \widehat\ell_{j,t}=\min\!\left(
    c_{\max},\frac{\ell_{j,t}}
    {\max\!\left(\sigma_{d_j}^{(\mu)},0.1\right)}
  \right),
  \label{eq:lsm}
\end{equation}
where $c_{\max}=5$ in the selected model. Scaling without mean centering preserves non-negativity. With the scale detached, tokens whose normalized loss exceeds $c_{\max}$ before clipping contribute zero direct gradient through the clipped term.

With fixed non-negative teacher weights $w_d$, the microbatch objective is
\begin{equation}
  \mathcal{L}_{\mu}(\theta)=
  \frac{\sum_{j\in\mathcal{B}_\mu}\sum_t
  m_{j,t}w_{d_j}\widehat\ell_{j,t}}
  {\sum_{j\in\mathcal{B}_\mu}\sum_t m_{j,t}}.
  \label{eq:mopd-total}
\end{equation}
The objective averages weighted losses over all valid response tokens (\texttt{token\_mean}). Domain sampling determines how often each teacher is selected, while $w_d$ multiplies that teacher's token losses. We fix $w_{\mathrm{IF}}=0.3$ and $w_d=1.0$ for Math, Code, and Agent throughout consolidation. The weighting configuration is discussed in \autoref{app:teacher-weighting}.

At each valid response position, the teacher supplies its Top-$K$ token identifiers and the corresponding log probabilities. The retained teacher outputs and local divergence computation scale as $O(TK)$ for response length $T$. The student forward pass still computes full-vocabulary logits.

\subsection{Efficient Execution for Dynamic MoE}
\label{sec:system}

\begin{figure}[t]
  \centering
  \includegraphics[width=\textwidth]{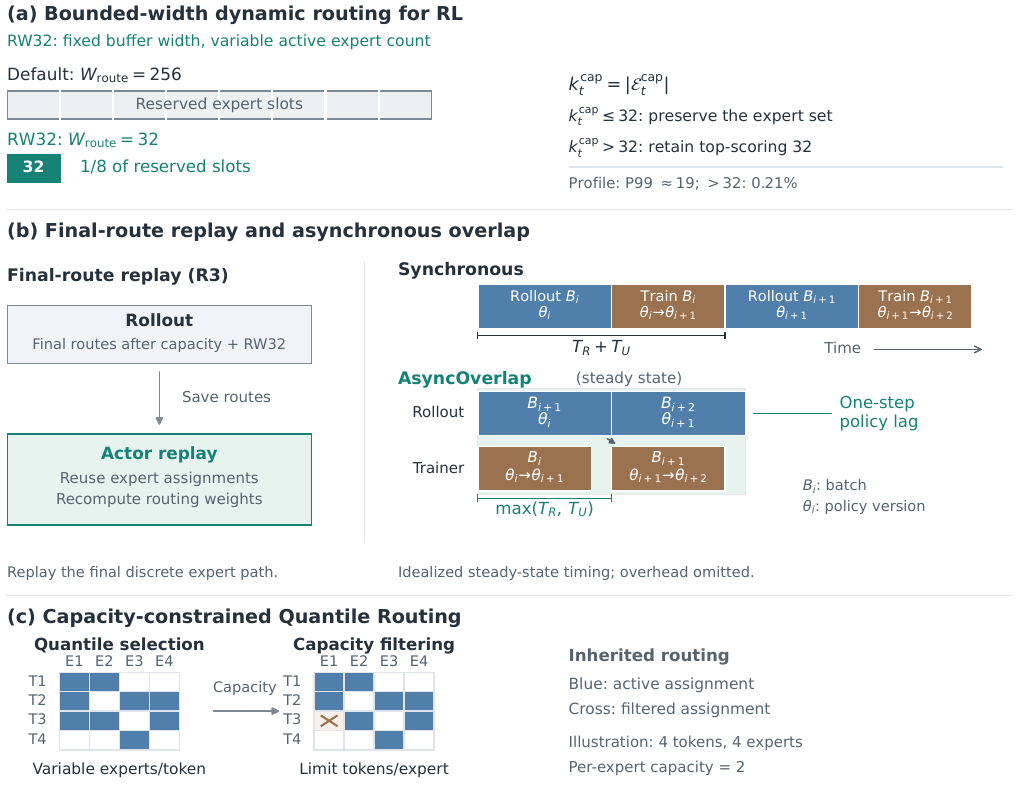}
  \caption{Dynamic-MoE execution during consolidation. (a) RW32 bounds post-capacity dispatch width at 32, preserving non-overflow expert assignments and retaining the highest-scoring 32 assignments on overflow. (b) R3-style final-route replay reuses the executed expert assignments and recomputes routing weights under the current actor parameters. AsyncOverlap overlaps rollout with actor updates using a one-step policy lag. (c) Base-model Quantile Routing selects a variable number of experts per token, followed by per-expert capacity filtering.}
  \label{fig:moe-execution-methods}
\end{figure}

The execution path applies capacity-constrained Quantile Routing, bounds the resulting dispatch width, and replays the final expert assignments during actor computation. These mechanisms are illustrated in \autoref{fig:moe-execution-methods}. Asynchronous overlap schedules rollout generation and actor updates concurrently.

\subsubsection{Capacity-Constrained Quantile Routing}
\label{sec:capacity-routing}

TuringLLM's Quantile Routing uses expert-specific score thresholds to select a token-adaptive number of experts under a controlled average budget~\cite{turingllm}. For a routing window $W$, per-expert capacity filtering retains at most $C(W)=\lceil\alpha |W|\bar{k}/E\rceil$ highest-scoring assignments among the selected candidates, where $\alpha$ is the capacity factor, $\bar{k}$ the target average expert count, and $E$ the number of experts. We retain this base-model routing rule throughout post-training (\autoref{fig:moe-execution-methods}c).

\subsubsection{Post-Capacity Dispatch-Width Bounding}

The default rollout representation reserves $E=256$ expert slots per token. In the measured routing profile, the 99th percentile of active expert counts is approximately 19, and 0.21\% of token routes exceed 32 experts. Based on this profile, we set $W_{\mathrm{route}}=32$ after capacity filtering. Let $A^{\mathrm{cap}}$ denote the post-capacity assignment map, $s_{t,e}$ the routing score, and $\mathcal{E}^{\mathrm{cap}}_t=\{e:A^{\mathrm{cap}}_{t,e}=1\}$. The executed expert set is
\begin{equation}
  \mathcal{E}^{\mathrm{roll}}_t=
  \begin{cases}
    \mathcal{E}^{\mathrm{cap}}_t,
      & |\mathcal{E}^{\mathrm{cap}}_t|\leq W_{\mathrm{route}},\\
    \operatorname{TopK}_{e\in\mathcal{E}^{\mathrm{cap}}_t}
      (s_{t,e},W_{\mathrm{route}}),
      & |\mathcal{E}^{\mathrm{cap}}_t|>W_{\mathrm{route}}.
  \end{cases}
\end{equation}
RW32 preserves post-capacity expert assignments for routes containing at most 32 experts and retains the highest-scoring 32 assignments for wider routes (\autoref{fig:moe-execution-methods}a). This reduces the width-dependent route representation to one eighth of its default size. Here, $W_{\mathrm{route}}$ denotes expert dispatch width; $K$ in C-MOPD denotes the teacher-local vocabulary support size.

\subsubsection{Final-Route Replay and Asynchronous Overlap}

Capacity filtering depends on the tokens in each execution window. Because rollout and actor windows may differ, independently repeating this filtering can produce different expert assignments. Following R3~\cite{r3}, we replay the final assignment map $A^{\mathrm{roll}}$ obtained after capacity filtering and width bounding. The actor reuses these discrete assignments and recomputes continuous routing weights under its current parameters,
\begin{equation}
  A^{\mathrm{actor}}=A^{\mathrm{roll}},\qquad
  p^{\mathrm{actor}}_{t,e}=A^{\mathrm{roll}}_{t,e}
  \sigma(z^\theta_{t,e})\gamma,
\end{equation}
where $\gamma$ is an optional routing scale.

AsyncOverlap runs rollout generation concurrently with actor updates (\autoref{fig:moe-execution-methods}b). Trajectories are generated by the preceding student policy version and evaluated by the current actor under the replayed expert assignments, introducing a one-step policy lag.

% !TeX root = ../main.tex
\section{Experiments}
\label{sec:experiments}

\subsection{Experimental Setup}

\modelname has 20B total parameters, approximately 2B active parameters per token, and a native 128K context window~\cite{turingllm}. Post-training retains its architecture and dynamic Quantile Routing. Unless stated otherwise, C-MOPD denotes the selected $K=16$, $c_{\max}=5$ checkpoint. Comparisons with the shared SFT checkpoint evaluate the complete specialist-training and consolidation pipeline; the calibration components are assessed in \autoref{sec:calibration-analysis}.

We use a common generative evaluation pipeline based on OpenCompass task definitions~\cite{opencompass}. The suite covers knowledge, reasoning, mathematics, code, verifiable instruction following, and function calling~\cite{hendrycks2021mmlu,li2024cmmlu,huang2023ceval,wang2024mmlupro,rein2024gpqa,suzgun2023bbh,cobbe2021gsm8k,hendrycks2021math,chen2021humaneval,austin2021mbpp,zhou2023ifeval,bfcl}. All Qwen3-8B reference scores in this section were reproduced using the same pipeline as the corresponding Turing checkpoints. Prompt mode, decoding settings, and response-length limits are matched within each comparison unless stated otherwise.

Direct-response prompts insert an empty \texttt{<think></think>} pair, while reasoning-enabled prompts leave the reasoning channel open. BFCL v4 uses reasoning-enabled evaluation and is labeled ``Think'' in the tables. Training budgets and data mixtures are reported in \autoref{app:training-recipe}; benchmark prompting and scoring protocols are summarized in \autoref{app:evaluation-protocol}.

Differences in benchmark scores are expressed in percentage points; normalized integration scores and their differences are dimensionless. Bold marks the best reported value within each comparison group.

\subsection{Multi-Domain Capability Integration}
\label{sec:capability-integration}

Following MOPD~\cite{mopd}, we measure the fraction of each specialist's improvement over SFT recovered by the consolidated student. For each checkpoint, $s_d$ is the unweighted mean of benchmark scores within domain $d$, using the same benchmark composition for the student and its SFT and specialist references. We normalize this domain mean using the shared SFT checkpoint and the corresponding specialist, then average the recovery ratios across $n=4$ domains with equal weights:

\begin{equation}
  \widetilde{s}_d=
  \frac{s_d-s_d^{\mathrm{SFT}}}
       {s_d^{\mathrm{teacher}}-s_d^{\mathrm{SFT}}},\qquad
  \mathrm{Integration}=\frac{1}{n}\sum_{d=1}^{n}\widetilde{s}_d.
  \label{eq:ablation-norm-score}
\end{equation}

Math includes GSM8K and MATH in direct-response mode, and GSM8K and MATH-500 in reasoning mode. Code includes HumanEval, MBPP, and MBPP+ in both modes. IF uses IFEval, and Agent uses BFCL v4 in reasoning mode. The SFT and specialist references map to 0 and 1, respectively; normalized values are not clipped.

\begin{table}[H]
  \centering
  \caption{Normalized specialist-gain recovery. The SFT and specialist rows define the normalization references; Macro is the unweighted mean of the four domain ratios.}
  \label{tab:primary-integration}
  \footnotesize
  \setlength{\tabcolsep}{6pt}
  \renewcommand{\arraystretch}{1.12}
  \begin{tabular}{@{}lrrrrr@{}}
    \toprule
    \textbf{Reference / model} & \textbf{Math} & \textbf{Code} & \textbf{IF} & \textbf{Agent} & \textbf{Macro} \\
    \midrule
    SFT reference & 0.0000 & 0.0000 & 0.0000 & 0.0000 & 0.0000 \\
    Specialist references & 1.0000 & 1.0000 & 1.0000 & 1.0000 & 1.0000 \\
    \midrule
    C-MOPD ($K=16$, $c_{\max}=5$) & 0.9407 & 1.0093 & 0.9016 & 1.1435 & 0.9988 \\
    \bottomrule
  \end{tabular}
\end{table}

The selected model achieves a macro integration score of 0.9988, with all four domain recovery ratios at or above 0.9016. Math and IF retain 94.07\% and 90.16\% of their specialists' improvements over SFT, while the Code and Agent domain means exceed their specialist references. \autoref{tab:stage-transfer} reports stage-wise benchmark scores, and \autoref{app:raw-ablation} provides the complete raw results underlying these normalized ratios.

\begin{table}[H]
  \centering
  \caption{Stage-wise scores on the benchmarks directly associated with each specialist. All rows use direct-response evaluation except BFCL v4, which uses reasoning-enabled evaluation.}
  \label{tab:stage-transfer}
  \footnotesize
  \setlength{\tabcolsep}{5pt}
  \renewcommand{\arraystretch}{1.12}
  \begin{tabular}{@{}lrrrrr@{}}
    \toprule
    \textbf{Metric} & \textbf{SFT} & \textbf{Specialist} & \textbf{C-MOPD} & \textbf{$\Delta$ vs. SFT} & \textbf{$\Delta$ vs. specialist} \\
    \midrule
    GSM8K & 90.07 & \textbf{91.96} & 91.58 & +1.51 & $-0.38$ \\
    MATH & 65.86 & \textbf{74.50} & 74.10 & +8.24 & $-0.40$ \\
    MBPP & 69.31 & 72.75 & \textbf{73.28} & +3.97 & +0.53 \\
    HumanEval & 79.88 & \textbf{86.59} & 84.15 & +4.27 & $-2.44$ \\
    IFEval & 58.41 & \textbf{80.96} & 78.74 & +20.33 & $-2.22$ \\
    BFCL v4 (Think) & 20.54 & 32.04 & \textbf{33.69} & +13.15 & +1.65 \\
    \bottomrule
  \end{tabular}
\end{table}

\FloatBarrier
\subsection{Calibration and Hyperparameter Ablations}
\label{sec:calibration-analysis}

\subsubsection{Loss-Scale Normalization and Clipping}

We compare three calibration variants at a fixed teacher-local support size of $K=32$ (\autoref{tab:calibration-fixed-k}). The full objective achieves a macro integration score of 0.9962, compared with 0.9724 without normalization and clipping and 0.9834 with normalization but without clipping. Relative to both removal variants, the full objective scores higher in Code, IF, and Agent and lower in Math.

\begin{table}[H]
  \centering
  \caption{Calibration ablations at $K=32$, with sampled-token MOPD as a separate method reference. \textit{No calibration} removes normalization and clipping; \textit{No clipping} retains normalization.}
  \label{tab:calibration-fixed-k}
  \footnotesize
  \setlength{\tabcolsep}{4pt}
  \renewcommand{\arraystretch}{1.12}
  \begin{tabular}{@{}llrrrrr@{}}
    \toprule
    \textbf{Objective} & \textbf{Variant} & \textbf{Math} & \textbf{Code} & \textbf{IF} & \textbf{Agent} & \textbf{Macro} \\
    \midrule
    \multicolumn{2}{l}{Sampled-token MOPD} & 0.7747 & 0.7830 & \textbf{0.9836} & 1.1417 & 0.9208 \\
    \midrule
    \multirow{3}{*}{Top-$K$ ($K=32$)} & No calibration & \textbf{0.8938} & 0.9142 & 0.9756 & 1.1061 & 0.9724 \\
    & No clipping & 0.8736 & 0.9356 & 0.9592 & 1.1652 & 0.9834 \\
    & Full C-MOPD & 0.8298 & \textbf{0.9514} & \textbf{0.9836} & \textbf{1.2200} & \textbf{0.9962} \\
    \bottomrule
  \end{tabular}
\end{table}

Sampled-token MOPD is included as a separate method reference; it differs in both the matching objective and the calibration configuration.

\subsubsection{Support Size and Clipping Threshold}

At $c_{\max}=5$, $K=16$ achieves the highest observed macro score of 0.9988, closely followed by $K=32$ at 0.9962 (\autoref{tab:hyper-sensitivity}). Compared with $K=32$, $K=16$ also yields a higher minimum domain recovery ratio, 0.9016 versus 0.8298, while halving the retained teacher entries per token. At $K=32$, $c_{\max}=5$ achieves the highest macro score among the tested thresholds, compared with 0.9181 for $c_{\max}=2$ and 0.8829 for $c_{\max}=10$.

\begin{table}[H]
  \centering
  \caption{Normalized integration scores for two sweeps: support size $K$ with $c_{\max}=5$ fixed, and clipping threshold $c_{\max}$ with $K=32$ fixed.}
  \label{tab:hyper-sensitivity}
  \footnotesize
  \setlength{\tabcolsep}{3.5pt}
  \renewcommand{\arraystretch}{1.12}
  \begin{tabular}{@{}llrrrrr@{}}
    \toprule
    \textbf{Sweep} & \textbf{Value} & \textbf{Math} & \textbf{Code} & \textbf{IF} & \textbf{Agent} & \textbf{Macro} \\
    \midrule
    \multirow{4}{*}{Support size} & $K=8$ & 0.7846 & 0.9661 & 0.9264 & 1.1870 & 0.9660 \\
    & $K=16$ & \textbf{0.9407} & \textbf{1.0093} & 0.9016 & 1.1435 & \textbf{0.9988} \\
    & $K=32$ & 0.8298 & 0.9514 & \textbf{0.9836} & \textbf{1.2200} & 0.9962 \\
    & $K=64$ & 0.8250 & 0.8226 & 0.9016 & 1.1809 & 0.9325 \\
    \midrule
    \multirow{3}{*}{Clipping threshold} & $c_{\max}=2$ & 0.7949 & 0.9403 & 0.8772 & 1.0600 & 0.9181 \\
    & $c_{\max}=5$ & \textbf{0.8298} & \textbf{0.9514} & \textbf{0.9836} & \textbf{1.2200} & \textbf{0.9962} \\
    & $c_{\max}=10$ & 0.7833 & 0.7651 & 0.8772 & 1.1061 & 0.8829 \\
    \bottomrule
  \end{tabular}
\end{table}

\FloatBarrier
\subsection{Broader Capability Evaluation}
\label{sec:broad-evaluation}

Relative to SFT, C-MOPD improves 10 of the 12 reported metrics (\autoref{tab:broad-results}). The largest gains are on IFEval (+20.33 points), BFCL v4 (+13.15), and MATH (+8.24). CMMLU remains nearly unchanged ($-0.01$), while GPQA Diamond decreases by 5.05 points, from 40.40 to 35.35.

\begin{table}[H]
  \centering
  \caption{Broader benchmark results. The first eleven rows use direct-response evaluation; BFCL v4 uses reasoning-enabled evaluation. $\Delta$ denotes the change from Turing SFT.}
  \label{tab:broad-results}
  \footnotesize
  \setlength{\tabcolsep}{4pt}
  \renewcommand{\arraystretch}{1.10}
  \begin{tabular*}{\textwidth}{@{\extracolsep{\fill}}lrrrr@{}}
    \toprule
    \textbf{Metric} & \textbf{Qwen3-8B} & \textbf{Turing SFT} & \textbf{C-MOPD} & \textbf{$\Delta$ vs. SFT} \\
    \midrule
    \multicolumn{5}{c}{\textit{General Knowledge and Reasoning Tasks}} \\
    \midrule
    MMLU & 74.23 & 76.59 & \textbf{78.10} & +1.51 \\
    CMMLU & 76.81 & \textbf{83.49} & 83.48 & $-0.01$ \\
    C-Eval & 67.74 & 80.48 & \textbf{81.07} & +0.59 \\
    MMLU-Pro & 25.71 & 45.75 & \textbf{48.53} & +2.78 \\
    BBH & \textbf{80.47} & 72.10 & 74.25 & +2.15 \\
    GPQA Diamond & 37.88 & \textbf{40.40} & 35.35 & $-5.05$ \\
    \midrule
    \multicolumn{5}{c}{\textit{Mathematics Tasks}} \\
    \midrule
    GSM8K & 88.32 & 90.07 & \textbf{91.58} & +1.51 \\
    MATH & \textbf{79.24} & 65.86 & 74.10 & +8.24 \\
    \midrule
    \multicolumn{5}{c}{\textit{Code Tasks}} \\
    \midrule
    MBPP & \textbf{75.13} & 69.31 & 73.28 & +3.97 \\
    HumanEval & 80.49 & 79.88 & \textbf{84.15} & +4.27 \\
    \midrule
    \multicolumn{5}{c}{\textit{Instruction-Following Task}} \\
    \midrule
    IFEval & \textbf{83.18} & 58.41 & 78.74 & +20.33 \\
    \midrule
    \multicolumn{5}{c}{\textit{Agent Task}} \\
    \midrule
    BFCL v4 (Think) & \textbf{38.64} & 20.54 & 33.69 & +13.15 \\
    \bottomrule
  \end{tabular*}
\end{table}

With reasoning enabled and an 8K response cap, C-MOPD improves over SFT on all four mathematics and code benchmarks (\autoref{tab:think-results}). Its unweighted mean score is 85.92, compared with 74.21 for SFT and 80.33 for Qwen3-8B.

\begin{table}[H]
  \centering
  \caption{Reasoning-enabled mathematics and code evaluation with an 8K response cap for all models.}
  \label{tab:think-results}
  \footnotesize
  \setlength{\tabcolsep}{5pt}
  \renewcommand{\arraystretch}{1.12}
  \begin{tabular}{@{}lrrrrr@{}}
    \toprule
    \textbf{Metric} & \textbf{Qwen3-8B} & \textbf{Turing SFT} & \textbf{C-MOPD} & \textbf{$\Delta$ vs. SFT} & \textbf{$\Delta$ vs. Qwen3} \\
    \midrule
    GSM8K & 92.72 & 92.95 & \textbf{95.68} & +2.73 & +2.96 \\
    MATH-500 & 80.80 & 73.00 & \textbf{82.40} & +9.40 & +1.60 \\
    MBPP & 62.43 & 55.29 & \textbf{77.78} & +22.49 & +15.35 \\
    HumanEval & 85.37 & 75.61 & \textbf{87.80} & +12.19 & +2.43 \\
    \bottomrule
  \end{tabular}
\end{table}

\subsubsection{Output-Length Budgets}

\autoref{tab:length-ablation} reports output-cap hit rates, defined as the percentage of responses reaching the configured generation limit. At 8K, C-MOPD has fewer cap hits than Qwen3-8B on all four tasks, alongside the higher accuracies in \autoref{tab:think-results}. Across 2,382 responses, 22 C-MOPD outputs (0.92\%) reach the cap, compared with 197 (8.27\%) for Qwen3-8B.

Increasing the Qwen3-8B cap to 32K raises its unweighted mean accuracy from 80.33 to 84.64, compared with 85.92 for C-MOPD at 8K. On MATH-500, Qwen3-8B at 32K scores 95.00, exceeding C-MOPD's 82.40 at 8K.

\begin{table}[H]
  \centering
  \caption{Output-cap hit rates under reasoning-enabled prompts and Qwen3-8B accuracy at 32K. A cap hit is a response that reaches the configured token limit. Corresponding 8K accuracies are reported in \autoref{tab:think-results}. Bold marks the lowest cap-hit rate for each task.}
  \label{tab:length-ablation}
  \footnotesize
  \setlength{\tabcolsep}{5pt}
  \renewcommand{\arraystretch}{1.12}
  \begin{tabular}{@{}lrrrr@{}}
    \toprule
    & \multicolumn{3}{c}{\textbf{Cap hits (\%)}} & \textbf{Accuracy (\%)} \\
    \cmidrule(lr){2-4}\cmidrule(l){5-5}
    \textbf{Metric} & \textbf{C-MOPD 8K} & \textbf{Qwen3 8K} & \textbf{Qwen3 32K} & \textbf{Qwen3 32K} \\
    \midrule
    GSM8K     & \textbf{0.08} & 1.74 & 0.30 & 92.95 \\
    MATH-500  & 3.60 & 19.00 & \textbf{2.60} & 95.00 \\
    HumanEval & \textbf{0.00} & 12.20 & 10.37 & 86.59 \\
    MBPP      & \textbf{0.75} & 14.79 & 9.52 & 64.02 \\
    \bottomrule
  \end{tabular}
\end{table}

\FloatBarrier
\subsection{Dynamic-MoE Consolidation Efficiency}
\label{sec:execution-profile}

We compare serial \emph{Consolidate}, one-step \emph{AsyncOverlap}, and \emph{AsyncOverlapRW32}, which combines asynchronous scheduling with a post-capacity expert-dispatch width limit of 32. Measurements cover the consolidation stage and are averaged over a common profiling interval (\autoref{tab:eff-profile}).

The combined configuration reduces step time from 398.7 to 139.3 seconds, yielding an observed 2.86$\times$ speedup. AsyncOverlap alone reduces trainer waiting but increases rollout time, resulting in a step time of 330.1 seconds. RW32 is profiled only with asynchronous overlap.

\begin{table}[H]
  \centering
  \caption{Average consolidation-system measurements over the common profiling interval.}
  \label{tab:eff-profile}
  \footnotesize
  \setlength{\tabcolsep}{5pt}
  \renewcommand{\arraystretch}{1.12}
  \begin{tabular}{@{}lrrr@{}}
    \toprule
    \textbf{Metric} & \textbf{Consolidate} & \textbf{AsyncOverlap} & \textbf{AsyncOverlapRW32} \\
    \midrule
    Step time (s) & 398.7 & 330.1 & \textbf{139.3} \\
    Trainer wait (s) & 265.0 & 195.8 & \textbf{18.3} \\
    Trainer compute (s) & 133.7 & 134.3 & \textbf{121.0} \\
    Rollout time (s) & 242.9 & 312.1 & \textbf{93.9} \\
    Rollout throughput (tok/GPU/s) & 183.9 & 143.1 & \textbf{478.4} \\
    Actor throughput (tok/s) & 26,966 & 26,451 & \textbf{31,165} \\
    Actor TFLOPS & 7.51 & 7.37 & \textbf{8.68} \\
    \bottomrule
  \end{tabular}
\end{table}

\begin{figure}[H]
  \centering
  \includegraphics[width=0.8\textwidth]{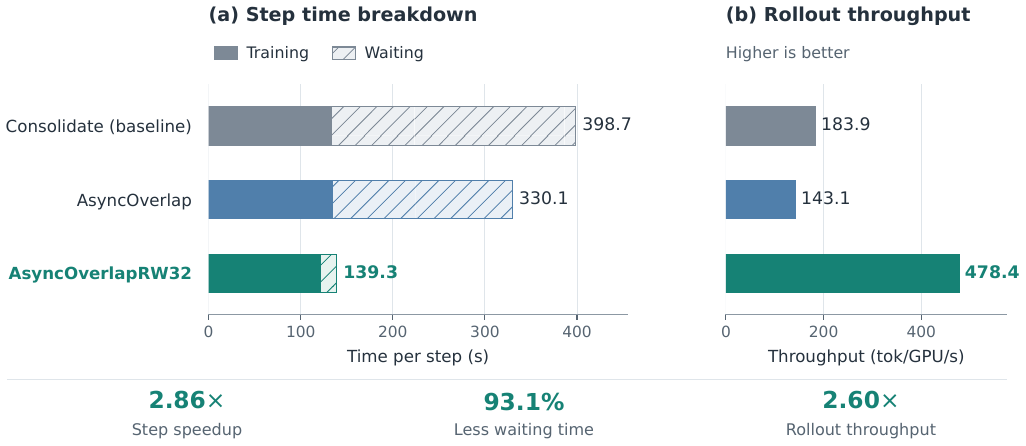}
  \caption{Consolidation-system profile. (a) Mean step time decomposed into trainer compute and waiting. (b) Rollout throughput in tokens per GPU per second. Summary values compare AsyncOverlapRW32 with serial Consolidate. Rollout generation may overlap trainer phases.}
  \label{fig:system-efficiency}
\end{figure}

Code benchmark scores at each configuration's selected checkpoint are reported in \autoref{app:execution-quality}. These checkpoints occur at different training steps, so they provide a descriptive comparison of execution-checkpoint quality.

% !TeX root = ../main.tex
\section{Conclusion}

We presented an end-to-end post-training framework that consolidates four independently trained domain specialists into a single dynamic-MoE student. C-MOPD calibrates multi-teacher on-policy distillation through teacher-wise loss-scale normalization and token-level clipping. The selected model achieves a normalized four-domain integration score of 0.9988 and retains at least 90.16\% of the corresponding specialist's measured improvement over the shared SFT baseline in every domain. At a fixed support size of $K=32$, the full objective achieves a higher macro integration score than variants with calibration components removed. The complete specialist-training and consolidation pipeline also improves over SFT on all four reasoning-enabled mathematics and code benchmarks and on ten of twelve metrics in the broader evaluation suite.

On PPU hardware, the execution workflow combines asynchronous rollout--training overlap, bounded-width expert dispatch after capacity filtering, and final-route replay. This combined configuration reduces the profiled average consolidation step time from 398.7 to 139.3 seconds, yielding an observed 2.86$\times$ speedup over serial execution.
% !TeX root = ../main.tex
\section{Contributors}
We sincerely thank every member of the team for their dedication and contributions to the Turing post-training project. This work brings together data construction, supervised fine-tuning, domain-specific reinforcement learning, calibrated multi-teacher on-policy distillation, MoE route replay and systems optimization, evaluation, and technical-report preparation. The results reflect a close collaboration across algorithm research, data and reward engineering, distributed training infrastructure, and model evaluation.

\textbf{Advisors}: Hang Zhang, Honggou Yang, Xianming Liu

\textbf{Project Lead}: Qiman Wu

\textbf{Contributors}:
Yue He\textsuperscript{*},
Jiahui Hu\textsuperscript{*},
Shaman Tang,
Hua Zhou,
Menglin Li,
Yizhao Wang,
Yuheng Zhang,
Da Zhu,
Zhou Wang,
Anhua Liu,
Hanlin Chen,
Mingyuan Wang,
Rui Xin,
Yuhua Wei,
Shuang Su,
Siqi Liu,
Kai Yang,
Weichao Huang.

\textsuperscript{*}Core contribution.

% \needsupplement{Confirm the public author order, corresponding author, and whether the current team attribution should remain on the title page.}

{
\begingroup
\sloppy
\small
\bibliographystyle{turing_unsrtnat}
\bibliography{main}
\endgroup
}

\clearpage
\appendix
\section*{Appendix}
% !TeX root = ../main.tex
\section{Training Recipe Details}
\label{app:training-recipe}

\autoref{tab:training-budget} summarizes the training budgets for shared SFT, specialist RL, and C-MOPD consolidation.

\begin{table}[H]
  \centering
  \caption{Post-training stages and data budgets. Pool sizes are rounded. RL and C-MOPD batch sizes distinguish sampled prompts from rollouts per prompt; epochs denote approximate passes through the corresponding data pool.}
  \label{tab:training-budget}
  \footnotesize
  \setlength{\tabcolsep}{5pt}
  \renewcommand{\arraystretch}{1.12}
  \begin{tabularx}{0.92\textwidth}{@{}lrrYr@{}}
    \toprule
    \textbf{Stage} & \textbf{Steps} & \textbf{Batch / step} & \textbf{Training pool} & \textbf{Epochs} \\
    \midrule
    SFT & 10,000 & 512 packed rows & 5.27M samples & $\sim$1.0 \\
    Code RL & 240 & 128 prompts $\times$ 16 rollouts & 16K prompts & $\sim$2.0 \\
    IF RL & 480 & 128 prompts $\times$ 16 rollouts & 16.2K prompts & $\sim$4.0 \\
    Math RL & 550 & 128 prompts $\times$ 16 rollouts & 35.5K prompts & $\sim$2.0 \\
    Agent RL & 512 & 64 prompts $\times$ 16 rollouts & 8.5K prompts & $\sim$4.0 \\
    C-MOPD & 400 & 128 prompts $\times$ 1 rollout & 51.5K prompts & $\sim$1.0 \\
    \bottomrule
  \end{tabularx}
\end{table}

\subsection{Supervised Data and Response Modes}

The shared SFT mixture contains approximately 5.27M examples (\autoref{tab:sft-mixture}). Qwen3.5-Plus and DeepSeek-V4-Flash provide mathematics and code supervision~\cite{qwen35plus,deepseekv4flash}. For selected source prompts, the current Turing checkpoint generates ten candidate responses. Incorrect responses are retained for teacher correction, which produces corrected SFT targets. Reasoning-enabled targets preserve the reasoning trace, while direct-response targets retain only the final response and use an empty \texttt{<think></think>} assistant prefix. Across the full mixture, approximately 53\% of examples use the reasoning-enabled form and 47\% use the direct-response form. Within mathematics and code distillation, the reasoning-enabled share is 85\%.

\begin{table}[H]
  \centering
  \caption{Composition of the SFT mixture. Counts are rounded to two decimal places in millions; shares use unrounded counts.}
  \label{tab:sft-mixture}
  \footnotesize
  \setlength{\tabcolsep}{5pt}
  \renewcommand{\arraystretch}{1.12}
  \begin{tabularx}{0.82\textwidth}{@{}Yrr@{}}
    \toprule
    \textbf{Data category} & \textbf{Examples} & \textbf{Share} \\
    \midrule
    Mathematics and code distillation & 2.69M & 51.0\% \\
    Agent tool use & 1.51M & 28.7\% \\
    General and instruction following (including ICL) & 0.67M & 12.7\% \\
    Long context and dialogue & 0.40M & 7.6\% \\
    \midrule
    \textbf{Total} & \textbf{5.27M} & \textbf{100\%} \\
    \bottomrule
  \end{tabularx}
\end{table}

\subsection{Capability-Boundary Filtering}

During iterative data construction, the current SFT checkpoint generates $n\geq4$ responses at temperature 1.0 for each prompt $x$. A task verifier assigns $r_i=1$ to a correct response and $r_i=0$ otherwise. The empirical pass rate is
\begin{equation}
  \widehat p(x)=\frac{1}{n}\sum_{i=1}^{n}r_i.
\end{equation}
We retain prompts satisfying
\begin{equation}
  \frac{2}{n}\leq\widehat p(x)\leq\frac{n-2}{n}.
  \label{eq:rejection-sampling-filter}
\end{equation}
This offline filter retains prompts with at least two verified successes and at least two failures among the sampled responses, selecting prompts with mixed observed outcomes under the current checkpoint.

% Author record: specify the sampling count n for this filter. Do not infer it
% from the ten candidate responses used in the separate teacher-correction step.

\subsection{Multiple-Choice Answer-Format Randomization}

For a multiple-choice item $(q,C,\ell)$, let $q$ denote the question, $C$ the candidate list, and $\ell$ the correct candidate in the original order. We sample a prompt template $\tau$, an answer format $f$, and a candidate permutation $\varpi$. The target map $g_{f,\varpi}(\ell)$ accounts for the selected format and shuffled candidate order. A deterministic parser $E_f$ returns $\bot$ when the response violates the format. The joint correctness-and-format reward is
\begin{equation}
  r_{\mathrm{mcq}}(y)=
  \mathbf{1}\!\left[E_f(y)\neq\bot\right]\,
  \mathbf{1}\!\left[E_f(y)=g_{f,\varpi}(\ell)\right].
\end{equation}
The format library contains 29 English and 29 Chinese variants, including alphabetic, numeric, Chinese-ordinal, and short-text forms. The reward is one only when the response follows the selected format and identifies the correct option.

\subsection{C-MOPD Domain Sampling and Loss Configuration}
\label{app:cmopd-sampling}

The consolidation pool contains 51,500 prompts: 15,000 for Math, 10,000 for Code, 16,500 for IF, and 10,000 for Agent. These counts describe the consolidation pool; the specialist RL pool sizes are listed separately in \autoref{tab:training-budget}. The domain counts define the prompt-sampling distribution
\[
  (\rho_{\mathrm{Math}},\rho_{\mathrm{Code}},
   \rho_{\mathrm{IF}},\rho_{\mathrm{Agent}})
  \approx(0.291,\,0.194,\,0.320,\,0.194).
\]
The prompt-domain sampling probabilities and teacher-loss multipliers remain fixed throughout the 400 consolidation steps, including 20 warmup steps and 380 subsequent steps. Valid response-token losses use the \texttt{token\_mean} reduction in \autoref{eq:mopd-total}. The fixed teacher-loss multipliers, applied after normalization and clipping, are given in \autoref{app:teacher-weighting}. The selected C-MOPD configuration uses $K=16$, a standard-deviation floor of 0.1, and $c_{\max}=5$; the ablations vary $K$ and $c_{\max}$ as specified in \autoref{sec:calibration-analysis}.

% Author record: identify the warmup schedule and the actual sampling procedure.

\section{Fixed Teacher Weights and Specialist Diagnostics}
\label{app:teacher-weighting}

\begin{figure}[H]
  \centering
  \includegraphics[width=\textwidth]{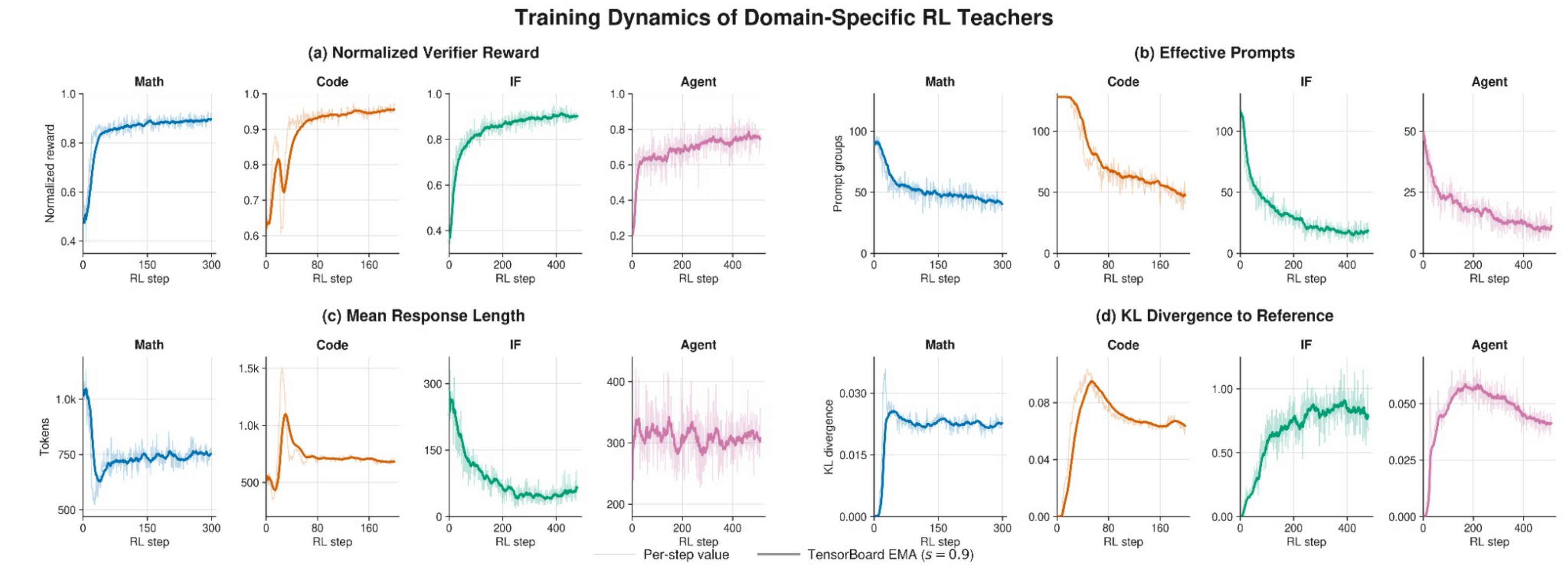}
  \caption{Specialist RL diagnostics over the displayed training intervals. Normalized verifier reward tracks task-specific rollout quality; the effective-prompt count records groups with non-degenerate relative-advantage signals; response length is measured in generated tokens; KL is measured against each specialist's RL reference policy rather than the C-MOPD student. Light lines show per-step values and dark lines show TensorBoard EMA curves ($s=0.9$).}
  \label{fig:expert-training-dynamics}
\end{figure}

The teacher-loss multipliers $w_d$ in \autoref{eq:mopd-total} are fixed before consolidation as $(w_{\mathrm{Math}},w_{\mathrm{Code}},w_{\mathrm{IF}},w_{\mathrm{Agent}})=(1.0,1.0,0.3,1.0)$ and remain unchanged throughout training. \autoref{fig:expert-training-dynamics} summarizes specialist RL behavior. These curves provide diagnostic context; the multipliers are fixed hyperparameters rather than values derived from the plotted statistics.

Reward and effective-prompt counts describe training behavior within each domain's verifier and sampling setup. The effective-prompt counts also depend on the per-step prompt budgets, which differ between Agent and the other domains (\autoref{tab:training-budget}). Mean response length records tokens generated during specialist RL.

During consolidation, $\rho_d$ controls prompt-domain sampling and $w_d$ multiplies the corresponding calibrated token losses. The scale statistic $\sigma_d^{(\mu)}$ rescales token losses, and the threshold $c_{\max}$ caps their normalized values. Under \texttt{token\_mean}, a domain's contribution also depends on its valid student-response token count. Scale calibration acts on loss values and does not normalize the number of contributing tokens.

% Author record: connect the plotted diagnostic intervals to the teacher runs
% and checkpoints used in consolidation. The Math panel ends near step 300,
% whereas the training-budget table reports 550 Math RL steps.

\section{Evaluation Protocol}
\label{app:evaluation-protocol}

\autoref{tab:eval-config} summarizes the benchmark splits, prompting settings, and scoring metrics. Few-shot and CoT settings describe benchmark prompting, while the direct-response and Think labels specify the model's response mode (\autoref{sec:experiments}).

\begin{table}[H]
  \centering
  \caption{Benchmark splits, prompting settings, and scoring metrics used in the reported evaluations.}
  \label{tab:eval-config}
  \footnotesize
  \setlength{\tabcolsep}{4pt}
  \renewcommand{\arraystretch}{1.12}
  \begin{tabularx}{\textwidth}{@{}lYlc@{}}
    \toprule
    \textbf{Benchmark} & \textbf{Split} & \textbf{Prompting} & \textbf{Metric} \\
    \midrule
    MMLU / CMMLU & Test & 5-shot & Acc. \\
    C-Eval & Validation & 5-shot & Acc. \\
    MMLU-Pro & Test & 5-shot CoT & Acc. \\
    GPQA Diamond & Diamond &  5-shot CoT & Acc. \\
    BBH & All 27 tasks & 3-shot CoT & EM \\
    GSM8K & Test & 4-shot CoT & EM \\
    MATH & Test (5,000 examples) & 4-shot & EM \\
    MATH-500 & Test (500 examples) & 0-shot & EM \\
    HumanEval / MBPP & Standard benchmark splits & 0-shot & Execution \\
    IFEval & Official set (541 prompts) & 0-shot & Prompt-level strict Acc. \\
    BFCL v4 & Official suite & Benchmark-specific think mode & Official score \\
    \bottomrule
  \end{tabularx}
\end{table}

The direct-response mathematics and code comparisons use the same task-specific caps for C-MOPD and Qwen3: 512 tokens for GSM8K, 2,048 for MATH, and 8,192 for HumanEval and MBPP. The dedicated reasoning-enabled comparison uses an 8K cap; Qwen3 at 32K is reported as a larger-budget reference.

% Author records required before release: exact model and teacher checkpoints;
% evaluation code and dataset revisions; generation settings and answer
% extractors; checkpoint-selection rule; and the Qwen3-8B revision.
% Confirm the stated GPQA Diamond 5-shot CoT setting against its run record.
% Add the actual MBPP+ split, prompting, test suite, scoring population and
% filtering rules; do not infer them from reported rounded accuracies.

\subsection{Base-Model Reference}

\autoref{tab:base-comparison} reports base-model results separately because the base and post-trained checkpoints use different prompting and response formats. This reference uses GPQA Main, whereas the post-training comparison in \autoref{tab:broad-results} uses GPQA Diamond.

% Author record: document the base-model prompting and generation settings.

\begin{table}[H]
  \centering
  \caption{Base-model results under the base-model prompting setup. GPQA uses the Main subset.}
  \label{tab:base-comparison}
  \footnotesize
  \setlength{\tabcolsep}{8pt}
  \begin{tabular}{@{}lrr@{}}
    \toprule
    \textbf{Metric} & \textbf{Qwen3-8B Base} & \textbf{Turing-20B-A2B Pretrain} \\
    \midrule
    MMLU & 78.92 & \textbf{79.83} \\
    CMMLU & 81.28 & \textbf{84.17} \\
    C-Eval & \textbf{83.06} & 81.76 \\
    MMLU-Pro & \textbf{55.35} & 55.17 \\
    BBH & \textbf{73.67} & 73.61 \\
    GPQA Main & 34.82 & \textbf{37.05} \\
    GSM8K & \textbf{90.07} & 74.53 \\
    MATH & 56.12 & \textbf{62.20} \\
    \bottomrule
  \end{tabular}
\end{table}

\section{Full Raw C-MOPD Results}
\label{app:raw-ablation}

\autoref{tab:ablation-raw} provides the raw scores underlying the normalized integration analysis. Sampled-token MOPD serves as a separate method reference. The three Top-$K$ calibration variants use $K=32$, while the selected C-MOPD model uses $K=16$ and $c_{\max}=5$. The teacher reference in each row is the corresponding domain specialist. Unmarked rows use direct-response evaluation, and rows marked Think use reasoning-enabled evaluation.

\autoref{tab:ablation-raw-sweeps} reports the support-size and clipping-threshold sweeps. Domain scores and the macro integration score are calculated from these raw results using \autoref{eq:ablation-norm-score}.

\begin{table}[H]
  \centering
  \caption{Raw benchmark scores (\%) for the reference checkpoints, sampled-token MOPD, three calibration variants at $K=32$, and the selected C-MOPD model. Teacher denotes the corresponding domain specialist. Bold marks the highest score in each row.}
  \label{tab:ablation-raw}
  \footnotesize
  \setlength{\tabcolsep}{1.7pt}
  \renewcommand{\arraystretch}{0.95}
  \begin{tabular*}{\textwidth}{@{\extracolsep{\fill}}lrrrrrrr@{}}
    \toprule
    & \multicolumn{2}{c}{\textbf{References}} & \textbf{Sampled} & \multicolumn{3}{c}{\textbf{Objective controls at $K=32$}} & \textbf{Selected} \\
    \cmidrule(lr){2-3}\cmidrule(lr){5-7}\cmidrule(l){8-8}
    \textbf{Metric} & \textbf{SFT} & \textbf{Teacher} & \textbf{MOPD} & \makecell{No\\calibration} & \makecell{No\\clipping} & \textbf{Full} & \makecell{$K=16$\\Full} \\
    \midrule
    \multicolumn{8}{c}{\textit{Mathematics Tasks}} \\
    \midrule
    GSM8K & 90.07 & \textbf{91.96} & 90.45 & 91.81 & 91.13 & 91.66 & 91.58 \\
    MATH & 65.86 & \textbf{74.50} & 70.32 & 73.28 & 73.14 & 73.34 & 74.10 \\
    GSM8K (Think) & 92.95 & \textbf{95.68} & 95.53 & 95.38 & 95.53 & 95.38 & \textbf{95.68} \\
    MATH-500 (Think) & 73.00 & 83.00 & \textbf{83.60} & 82.20 & 82.40 & 80.80 & 82.40 \\
    \midrule
    \multicolumn{8}{c}{\textit{Code Tasks}} \\
    \midrule
    HumanEval & 79.88 & \textbf{86.59} & 80.49 & 85.98 & 85.37 & 82.93 & 84.15 \\
    MBPP & 69.31 & 72.75 & 69.84 & 72.22 & \textbf{73.81} & \textbf{73.81} & 73.28 \\
    MBPP+ & 57.41 & 62.17 & 61.64 & 63.23 & 62.96 & 63.23 & \textbf{64.02} \\
    HumanEval (Think) & 75.61 & 86.59 & 85.37 & 85.98 & 85.37 & \textbf{87.80} & \textbf{87.80} \\
    MBPP (Think) & 55.29 & \textbf{78.04} & 76.46 & 75.66 & 76.46 & 76.46 & 77.78 \\
    MBPP+ (Think) & 48.68 & \textbf{66.67} & 64.55 & 64.02 & 64.55 & 65.34 & 66.40 \\
    \midrule
    \multicolumn{8}{c}{\textit{Instruction-Following Task}} \\
    \midrule
    IFEval & 58.41 & \textbf{80.96} & 80.59 & 80.41 & 80.04 & 80.59 & 78.74 \\
    \midrule
    \multicolumn{8}{c}{\textit{Agent Task}} \\
    \midrule
    BFCL v4 (Think) & 20.54 & 32.04 & 33.67 & 33.26 & 33.94 & \textbf{34.57} & 33.69 \\
    \bottomrule
  \end{tabular*}
\end{table}

\begin{table}[H]
  \centering
  \caption{Raw benchmark scores (\%) for the support-size and clipping-threshold sweeps. Bold marks the highest score within each row and sweep block.}
  \label{tab:ablation-raw-sweeps}
  \footnotesize
  \setlength{\tabcolsep}{1.7pt}
  \renewcommand{\arraystretch}{0.95}
  \begin{tabular*}{\textwidth}{@{\extracolsep{\fill}}lrrrrrrr@{}}
    \toprule
    & \multicolumn{4}{c}{\textbf{Support size, $c_{\max}=5$}} & \multicolumn{3}{c}{\textbf{Clipping, $K=32$}} \\
    \cmidrule(lr){2-5}\cmidrule(l){6-8}
    \textbf{Metric} & $K=8$ & $K=16$ & $K=32$ & $K=64$ & $c_{\max}=2$ & $c_{\max}=5$ & $c_{\max}=10$ \\
    \midrule
    \multicolumn{8}{c}{\textit{Mathematics Tasks}} \\
    \midrule
    GSM8K & 90.75 & 91.58 & \textbf{91.66} & 91.21 & 90.30 & \textbf{91.66} & 90.90 \\
    MATH & 73.90 & \textbf{74.10} & 73.34 & 73.22 & 73.22 & \textbf{73.34} & 73.30 \\
    GSM8K (Think) & \textbf{95.68} & \textbf{95.68} & 95.38 & 94.84 & \textbf{95.45} & 95.38 & 95.30 \\
    MATH-500 (Think) & 79.80 & \textbf{82.40} & 80.80 & 81.80 & \textbf{81.40} & 80.80 & 80.60 \\
    \midrule
    \multicolumn{8}{c}{\textit{Code Tasks}} \\
    \midrule
    HumanEval & 82.93 & \textbf{84.15} & 82.93 & 81.10 & \textbf{84.76} & 82.93 & 82.93 \\
    MBPP & \textbf{75.13} & 73.28 & 73.81 & 73.54 & 72.22 & \textbf{73.81} & 70.37 \\
    MBPP+ & 62.70 & \textbf{64.02} & 63.23 & 61.90 & 62.17 & \textbf{63.23} & 60.58 \\
    HumanEval (Think) & 87.20 & \textbf{87.80} & \textbf{87.80} & 84.76 & \textbf{88.41} & 87.80 & 85.98 \\
    MBPP (Think) & 77.25 & \textbf{77.78} & 76.46 & 75.40 & 75.93 & \textbf{76.46} & 74.34 \\
    MBPP+ (Think) & 65.34 & \textbf{66.40} & 65.34 & 64.29 & \textbf{65.34} & \textbf{65.34} & 62.96 \\
    \midrule
    \multicolumn{8}{c}{\textit{Instruction-Following Task}} \\
    \midrule
    IFEval & 79.30 & 78.74 & \textbf{80.59} & 78.74 & 78.19 & \textbf{80.59} & 78.19 \\
    \midrule
    \multicolumn{8}{c}{\textit{Agent Task}} \\
    \midrule
    BFCL v4 (Think) & 34.19 & 33.69 & \textbf{34.57} & 34.12 & 32.73 & \textbf{34.57} & 33.26 \\
    \bottomrule
  \end{tabular*}
\end{table}

\section{Execution Configuration Quality}
\label{app:execution-quality}

\autoref{tab:eff-quality-best-step} reports timing and code benchmark scores for the PPU execution configurations, providing a run-level view of their quality--efficiency trade-off. The code scores characterize the model states evaluated for each configuration; because these states come from different training steps, score differences should not be attributed to the execution configuration alone. Common-interval system averages are reported in \autoref{tab:eff-profile}.

% Author records: selected checkpoint IDs and training steps; checkpoint
% selection metric; code-evaluation prompt mode, output cap and scoring rules;
% timing window and wait-percentage denominator; PPU model and device count.

\begin{table}[H]
  \centering
  \caption{Timing and code benchmark scores for the PPU execution configurations.}
  \label{tab:eff-quality-best-step}
  \footnotesize
  \setlength{\tabcolsep}{3pt}
  \renewcommand{\arraystretch}{1.0}
  \begin{tabularx}{\textwidth}{@{}Xrrrrrr@{}}
    \toprule
    \textbf{Configuration} & \makecell{\textbf{Step time}\\\textbf{(s)}} & \textbf{Wait (\%)} & \textbf{Speedup} & \textbf{HumanEval} & \textbf{MBPP} & \textbf{MBPP+} \\
    \midrule
    Consolidate & 381.4 & 65.4 & 1.0$\times$ & 86.59 & \textbf{78.04} & \textbf{66.67} \\
    AsyncOverlap & 335.3 & 59.8 & 1.1$\times$ & 84.76 & 76.46 & 65.34 \\
    AsyncOverlapRW32 & \textbf{137.2} & \textbf{12.4} & \textbf{2.8$\times$} & \textbf{87.20} & 76.19 & 65.87 \\
    \bottomrule
  \end{tabularx}
\end{table}

\end{document}